\documentclass{llncs}
\usepackage{amsmath,graphicx,amssymb,wrapfig,hyperref}
\usepackage{subfig}

\newcommand{\bibfont}{\small}
\begin{document}
\title{A Statistical Model for Simultaneous Template Estimation, Bias Correction, and Registration of 3D Brain Images}




\author{Akshay Pai\inst{1,2}, Stefan Sommer\inst{1}, Lars Lau Raket\inst{3}, Line K\"{u}hnel\inst{1}, Sune Darkner\inst{1}, Lauge S{\o}rensen\inst{1,2}, Mads Nielsen\inst{1,2}}



\institute{DIKU, University of Copenhagen, Copenhagen, Denmark \and Biomediq A/S, Copenhagen, Denmark \and Lundbeck, Denmark}


\maketitle
\begin{abstract}
Template estimation plays a crucial role in computational anatomy since it provides reference frames for performing statistical analysis of the underlying anatomical population variability. While building models for template estimation, variability in sites and image acquisition protocols need to be accounted for. To account for such variability, we propose a generative template estimation model that makes simultaneous inference of both bias fields in individual images, deformations for image registration, and variance hyperparameters. In contrast, existing maximum a posterori based methods need to rely on either bias-invariant similarity measures or robust image normalization. Results on synthetic and real brain MRI images demonstrate the capability of the model to capture heterogeneity in intensities and provide a reliable template estimation from registration. 
\end{abstract}

\section{Introduction}
Brain template estimation is becoming increasingly important since it facilitates a variety of applications such as segmentation, registration, or providing common coordinate systems for statistical analysis of shape models for a given population. At the core of template estimation methods is image registration. In a statistical setting, conventional image registration methods are often approached from a Bayesian viewpoint where one maximizes the posterior given the image data and a regularizing prior~\cite{Joshi:2004cr}. To avoid choosing parameters for controlling the deformation (regularization) in an ad-hoc fashion, recent methods have employed Bayesian models where parameters are estimated in a data-driven fashion by treating them as latent random variables drawn from a distribution with a smooth covariance structure~\cite{Zhang:2013it,Allassonniere:2010ya}. These methods employ $L^2$ similarity measures that are fragile towards deviations in model assumption; for instance, bias fields. To achieve reasonable results under model deviations, strong penalization on the variation in deformation parameters can be imposed, or similarity measures that are invariant to bias fields (e.g. mutual information) can be used instead of the $L^2$ data terms.

In this paper, we propose a statistical mixed-effects model where deformation and spatially correlated variation in intensity are modeled as random effects. This way, effects from deformations and variations in intensities due to scanners can be separately handled, and hyperparameters can be estimated in a data-driven fashion using maximum likelihood. We perform simultaneous estimation and prediction in the model to avoid bias in the estimation that can result from treating warping as a preprocessing step~\cite{Raket:2014uf}.
In addition, we propose a different estimation procedure in comparison to existing Bayesian methods. While current methods use non-linear sampling for marginalizing over the latent variables, we propose to use successive linearization around predictions of the latent variables allowing estimation with linear mixed-effect theory. This paper is built on the methods proposed by Raket et. al,~\cite{Raket:2014uf} for analysis of 1D functional data. 

Current probabilistic template estimation methods do not model multi-scale behavior of the deformations. Although innumerable multi-scale deformation models have been proposed for image registration, their application in template estimation still needs maturity. To this end, we propose to model deformations at different scales as latent variables that are drawn from different distributions with different covariance structure. Concretely, we utilize the kernel bundle structure~\cite{Sommer:2012gu,Pai:2015di} to model velocity fields, and the multi-scale nature is interleaved in the covariance structure of the distribution of parameters at each kernel bundle level.

The contributions of the paper are as follows:

\begin{itemize}
\item{We demonstrate the utility of a computationally feasible class of non-linear mixed effect models in 3D brain template estimation.}
\item{We propose a model that handles effects from deformation and intensity variations separately and allows simultaneous estimation of template and variance hyperparameters and prediction of deformation and bias fields.}
\item{We propose an iterative linearization of the model in the non-linear random effects that enables efficient maximum likelihood estimation of variance parameters.}
\item{Finally, we propose the incorporation of scales in the deformation distribution via the kernel bundle representation.}
\end{itemize}

\section{Background}

A number of registration-based approaches to template estimation have been proposed where the central aspect addressed was the choice of target co-ordinate system. Initial approaches, which include the popular minimum deformation template~\cite{Kochunov:2001mo}, choose a random image as a target co-ordinate system, and is iteratively updated by registering other images to it. This approach has been shown to be significantly biased towards the choice of the random image~\cite{Rueckert:2003ib}. 

As an alternate approach, several papers have proposed the strategy of registering several images to a template, which is simultaneously estimated via an alternating optimization scheme~\cite{Vialard:2011rq,Zollei:2007ta}. In a probabilistic formulation, the image matching term can be viewed as a log likelihood and the regularization as a log prior. In this direction, recent methods like~\cite{Allassonniere:2010ya} have employed tools such as non-linear mixed effects models to deal with the population effects (template or the fixed effect) and individual effects (deformations or the random effect). Methods have moved towards simultaneously modeling the template and inferring parameters of the deformation such as the regularization factor with expectation-maximization~\cite{Zhang:2013it}. All the aforementioned methods rely on bias-invariant similarity measures or robust image normalization methods. To address intensity bias factors, Hromatka et. al,~\cite{Hromatka:2015px}., proposed to use a hierarchical Bayesian model for template estimation where two transformations are concatenated; one taking an individual image to an atlas of a site and the other that takes this warped image to the global atlas. However, this model takes into consideration very little about intra-site bias variations. 

\section{Statistical model}
Consider a population of images $I_i:\mathbb{R}^3 \rightarrow \mathbb{R}, i=1\dots k$ and let $\theta$ be a template of these images; both measured on a discrete grid $\Omega \in \mathbb{Z}^3$. The individual observed image may be then defined in terms of the fixed and random effects as
\begin{equation}
\label{observation}
I_i = \theta(\text{Exp}(v(w_i)) + x_i +\epsilon_i
\end{equation}
where the template $\theta$ is the fixed effect. More control over the smoothness of $\theta$ may be incorporated by specifying a parametric subspace for $\theta$. However, such constructions will not be entertained in this paper. The remaining effects are all random: The deformation $\text{Exp}(v(w_i))$ (Section~\ref{deformation}) is a random field controlled by latent random deformation parameters $w_i$. A key contribution of this paper is the incorporation of the random spatially correlated effect $x_i$ that models a bias field. Note that $x_i$ is defined in the frame of the individual image and not the template. Converse constructions are also possible. Finally, $\epsilon_i$ is the i.i.d noise. 

Following~\cite{Raket:2014uf} and assuming that $\theta$ is smooth, we can linearize~\eqref{observation} around deformation parameters $w_i^0$ resulting in the linear model
\begin{align*}
\label{linear}
&I_i\approx \theta^{w_i^0} + Z_i(w_i-w_i^0) + x_i+ \epsilon_i, \text{where,}\\
&\theta^{w_i^0} = \theta (\text{Exp}(v(w_i^0))),~Z_i = \nabla_{\bold x}{\theta(\text{Exp}(v(w_i^0)))}^T|_{\bold x=\text{Exp}(v(w_i^0))}J_w \text{Exp}(v(w_i^0)) \in \mathbb{R}^{n\times n_w}\ , \\
&w_i\sim\mathcal{N}(0,\sigma^2C_i),~C_i=\mathbb{I}_i\otimes C_0,~x_i\sim\mathcal{N}(0,\sigma^2S_i),~S_i=\mathbb{I}_i\otimes S_0\ ,\\
&\epsilon_i\sim\mathcal{N}(0,\sigma^2\mathbb{I}_i)\ .
\end{align*}
Here $\sigma^2$ is the noise variance, and the spatial covariance of the deformations and bias fields are controlled by the matrices $C$ and $S$. Note that $\nabla_{\bold x}$ denotes derivatives with respect to spatial coordinates. Also, note that deformations in the linear model are parameters of $w$ while the linearization point $w^0$ is iteratively optimized for during the estimation process. The model here is a single scale model where the covariances are constructed with only one scale of the kernel. Multi-scale version of this will be discussed in Section~\ref{deformation}.

The first step of the analysis is to estimate $\theta$ with an initial guess of linearization point $w$. $\theta$ is computed by back-warping the images with the velocity field as $
\theta^w_0 = \frac{1}{k} \sum_{i=1}^k I_i (\text{Exp}(-v(w_0))).$
This simplified formulation of the conditional maximum-likelihood is a result of Henderson's mixed-model~\cite{Henderson:1950yr} that simplifies when all observations are on a common grid. With this estimate of $\theta$, we estimate the variance parameters by minimizing the double negative log-likelihood of the linearized model
\begin{equation}
\label{likelihood}
\begin{split}
&\mathcal{L}(\sigma^2, S, C) = nk~\text{log}~\sigma^2 + \sum_{i=1}^k \text{log}~\text{det} V_i + 
\\&\quad\quad\quad\quad\quad\quad\frac{1}{\sigma^2} \sum_{i=1}^k (I_i - \theta^{w_0} + Z_i w^0)^TV_i^{-1}(I_i - \theta^{w_0} + Z_i w^0).
\end{split}
\end{equation}
where $V_i = Z_i^TC_iZ_i + S + \mathbb{I}_i$. Computing~\eqref{likelihood} directly is computationally intractable. This is because the dimensionality of $S,~V$ is $m^2$, where $m$ is the number of voxels in the image. We handle this by two assumptions: a) The support of the kernel used to construct the covariance matrix of the spatial correlation effect is sufficiently large that the inverse of the covariance matrix resembles a block-diagonal matrix, see e.g. assumption~\cite{Si:2014sh}; and b) the interaction between the blocks is negligible. This allows the likelihood to be approximated by an integral over smaller computationally tractable patches over the image. Therefore,~\eqref{likelihood} can be rewritten as:
\begin{equation}
\label{likelihood1}
\begin{split}
&\mathcal{L}(\sigma^2, S, C) \approx nk~\text{log}~\sigma^2 + \sum_{i=1}^k \sum_{j=1}^p \text{log}~\text{det} V_{i,j} + 
\\&\quad\quad\quad\quad\quad\quad \frac{1}{\sigma^2} \sum_{i=1}^k \sum_{j=1}^p (I_{i,j} - \theta^{w_0} + Z_{i,j} w^0)^TV_{i,j}^{-1}(I_{i,j} - \theta_j^{w_0} + Z_{i,j} w^0).
\end{split}
\end{equation}
where $V_{i,j} = Z_{i,j}^TC_iZ_{i,j} + S_j + \mathbb{I}_{j}$. Here $j = 1 \dots p$ is the patch number of the image with total $p$ patches. The estimation process is now repeated: Given the current estimate of the template $\theta$ and the variance estimates, the new deformation parameters $w^0$ for the linearization point are chosen as the most likely predictions in the original non-linear model~\eqref{observation}. That is, they are given by minimizing the negative log posterior of the deformation given the image data in model~\eqref{observation}:
\begin{equation}
\mathcal{P}(w) = \sum_{i=1}^k \sum_{j=1}^p (I_{i,j} - \theta_j^{w_i})^T (S_j + I_j)^{-1}(I_{i,j} - \theta_j^{w_i}) + w_i^TC_iw_i.
\end{equation}
To reconstruct the bias field for a given iteration, the conditional expectation of the spatially correlated effect $x_i$ given the data $I_i$ and the most likely deformation parameters $w_i^0$  are computed under the maximum likelihood estimates for the model parameters:
\begin{equation}
\label{bias}
E[x_i|w^0_i,I_i] =  S (S+I)^{-1}(I_{i} - \theta_j^{w_i})\ . 
\end{equation}
Note that the best linear unbiased predictor (BLUP)~\cite{Robinson:1991fl} of $w_i$ in the linearized model given the image data $I_i$ is realized by $E[w_i|I_i]$ which is computed as $(C_i^{-1}+Z_i^T(\mathbb{I}_i+S)^{-1}Z_i)^{-1}Z^T(\mathbb{I}_i+S)^{-1}(I_i-\theta^{w_i^0}+Zw^i_0)$.
\subsection{Covariance matrices}
A key aspect of this statistical model is the choice of covariance matrices for the spatially correlated effects and the deformation effects. Current Bayesian methods model the inverse of the covariance matrix directly by the means of an operator; typically of the Cauchy-Navier type which takes the form, $L = -\alpha \nabla + \beta$ where $\nabla$ is the Laplace operator. The parameter $\alpha$ here controls the smoothness of the covariance. In this paper, since we work with patches, it is more intuitive to model the covariance matrix directly. We model the covariance matrices for both the deformation parameters and the spatial correlation using Wendland kernels~\cite{Wendland:1995hb} that are compactly supported reproducing kernels. The covariance matrices are constructed using $C = \lambda^2 K(c_i, c_j)\ ,\ \lambda \in (0,\infty)$, $K (a,b)= r^4(4r+1),~r= (1-t,0)_{+},~t =\frac{||a-b||}{s}$, where $c_i$ are the kernel centers. 

For the spatially correlated effect ($S = \beta K$), a similar representation is chosen. However, we choose the parametric subspace to be the same size as that of the image i.e., a kernel is centered at every voxel as opposed to the much smaller subspace spanned by the deformation kernels. The amplitude of these covariance matrices are controlled by parameters $\lambda, \beta$. These parameters are estimated by optimizing the likelihood \eqref{likelihood}. The smoothness of the deformation is controlled by $C$. We here keep $C$ fixed though parameters of $C$, e.g. range and scale, can also be optimized for in \eqref{likelihood}.  

\subsection{Multi-scale deformation model}
\label{deformation}
We model deformation fields as path of diffeomorphisms generated by integrating stationary velocity fields (SVFs).
Let $G\subset\text{Diff}(\Omega)$ be a Lie subgroup of the group of diffeomorphic transformations $\varphi :\Omega \rightarrow \Omega$, and let $\phi: \Omega \times \mathbb{R} \rightarrow \Omega$ be a path in $G$. Let $V$ be the tangent space of $G$ at identity $\text{Id}$ containing velocity fields ${v}: \Omega\rightarrow\mathbb{R}^d$. In SVFs, the governing differential equation can be written as:
\begin{equation}
\label{svf}
\begin{split}
\frac{\partial{\phi(\bold{x},t)}}{\partial{t}} &= {v}(\phi(\bold{x},t)),~\varphi=\phi(\bold{x},1) =\text{Exp}({v}) \ .
\end{split}
\end{equation}
The final transformation $\phi$ is the Lie group exponential $\text{Exp}({v})$. The integration in \eqref{svf} can be numerically realized as an Euler integration. 
We use the kernel bundle framework~\cite{Sommer:2012gu,Pai:2015di} in this paper. In short, the concept of the space of velocity fields $V$ is extended to a family  $\hat{V}$ of spaces of velocity fields. The velocity fields are linear sums of individual kernels at $R$ levels
\begin{equation}
\label{form1}
{v}(\bold{x})=\sum_{m=1}^R  v_m=\sum_{m=1}^R\sum_{i}^{N_m} K_m( c_{i}^m, \bold x) w_{i}^m\ .
\end{equation}
Here $K$ is an interpolating Wendland kernel, $R$ is the total number of kernel bundle levels, $N_m$ is the number of kernels at each level and $c$ is center of each kernel at the kernel bundle level. The parameter $w^m$  is associated with the $m$th kernel bundle level, and we assume that
$w_{i}^m \sim \mathcal{N}(0,\sigma^2C^m)$,
where $C^m$ is the covariance matrix for each kernel bundle level with its distinct support and smoothness. 
\section{Experiments and Results}
We perform an evaluation on the MGH10 dataset\footnote{www.mindboggle.info}. The dataset contains 10 images each with the dimension of 182x218x182 and a voxel resolution of 1,1.33,1 mm. The images are initially co-registered rigidly. For bias recovery only a subset of 5 subjects is used.
\begin{figure}[h]
\centering
\minipage{0.2\textwidth}
  \includegraphics[width=\linewidth]{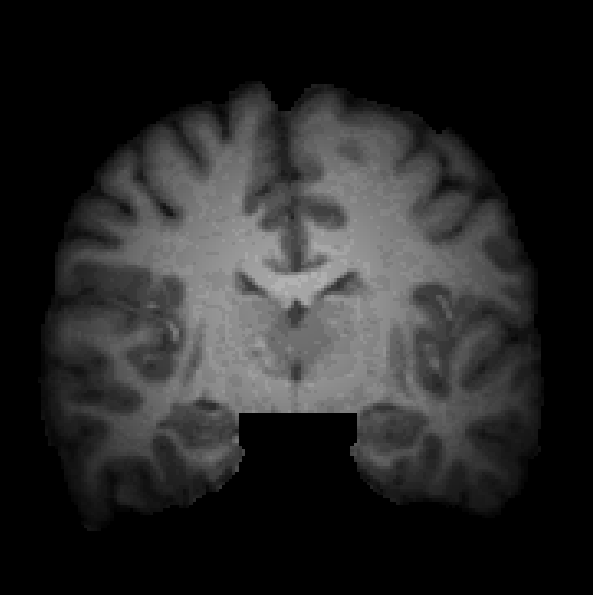}
\endminipage
\hspace{1cm}
\minipage{0.2\textwidth}
  \includegraphics[width=\linewidth]{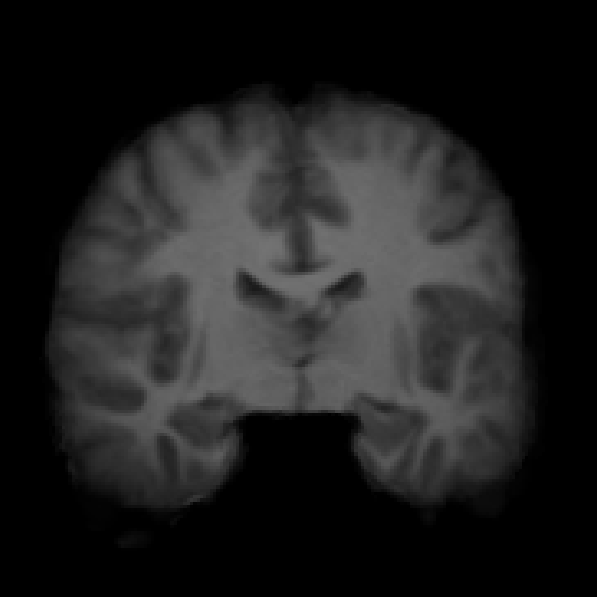}
\endminipage
\caption{An illustration of the image with bias field (left) and the recovered image (right).}
\label{fig:image3}
\end{figure}
One of the major challenges in the field of image template estimation is validation. Since the underlying geometry of the space of images is unknown, the definition of modes of populations is difficult. Therefore, the validation needs to be tied to specific applications of template estimation like image segmentation or even validation of the underlying image registrations. For the 2 experiments in this paper, we chose 3 kernel bundle levels each with a control point spacing of 20, 10, 6 mm respectively. The support of the kernel used to construct the covariance matrix for the deformations is fixed to 4 across the kernel bundle levels. The support of the spatial correlation variance kernel was set to 40. We perform 2 experiments to validate our template estimation method.
\subsection{Bias field recovery}
We select a subset of 5 images from the database and add artificial multiplicative bias to the image of form $\text{bias} = 1+\text{exp}(-\frac{\bold x^2 + \bold y^2 + \bold z^2}{2*30^2})*0.05.$. 
We then apply our statistical model to estimate the template and inspect whether the corrupted image can be recovered. As illustrated in Figure~\ref{fig:image3} the image recovered is free of the bias field illustrating the robustness of the method towards scanner-related artifacts. 
\subsection{Template estimation}
\begin{figure}
\centering
\minipage{0.25\textwidth}
  \includegraphics[width=5cm]{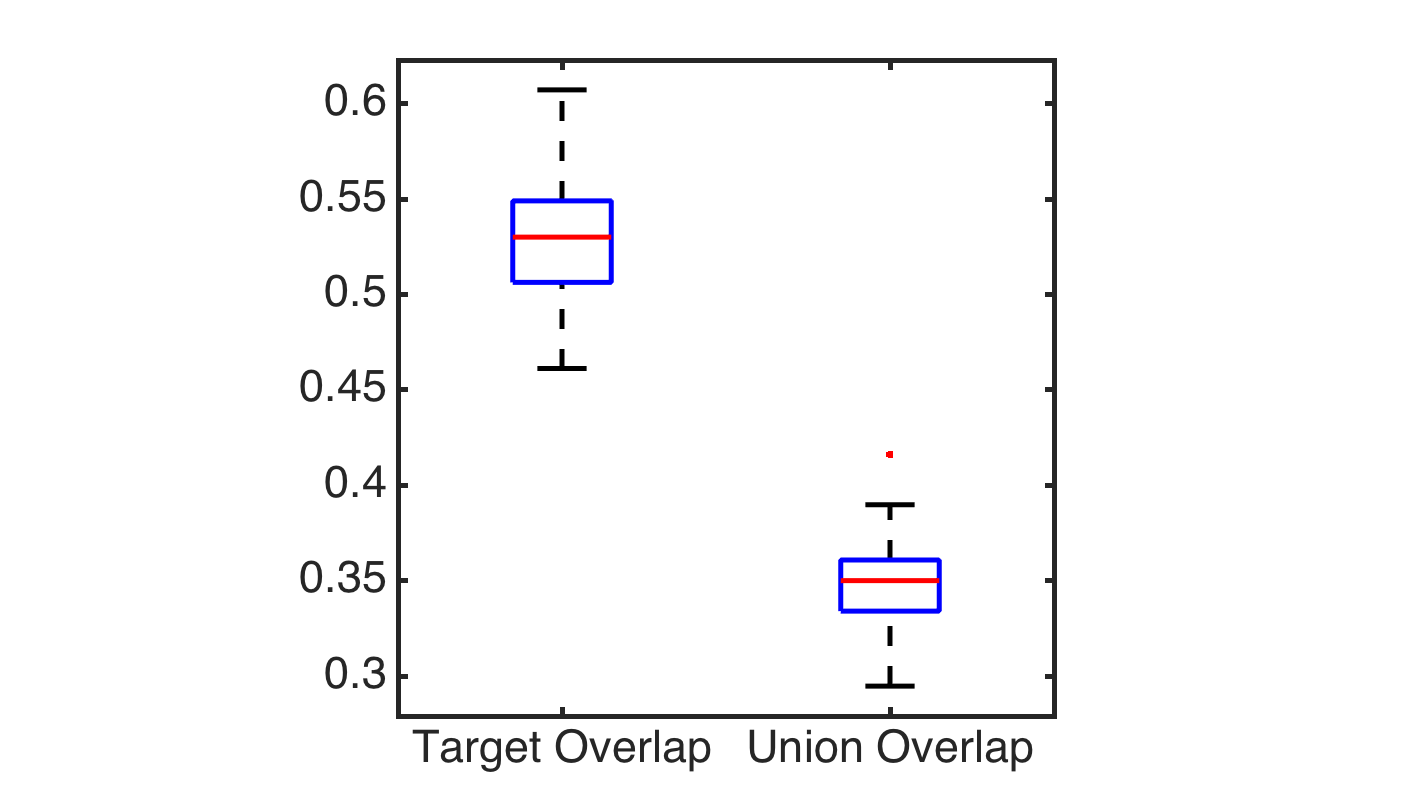}
\endminipage
\hspace{1cm}
\minipage{0.25\textwidth}
  \includegraphics[width=\linewidth]{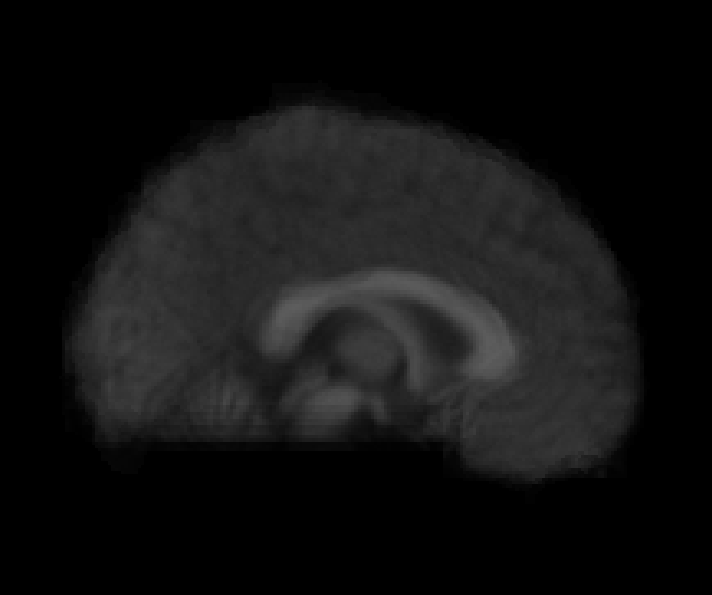}
\endminipage
\hspace{0.3cm}
\minipage{0.25\textwidth}
  \includegraphics[width=\linewidth]{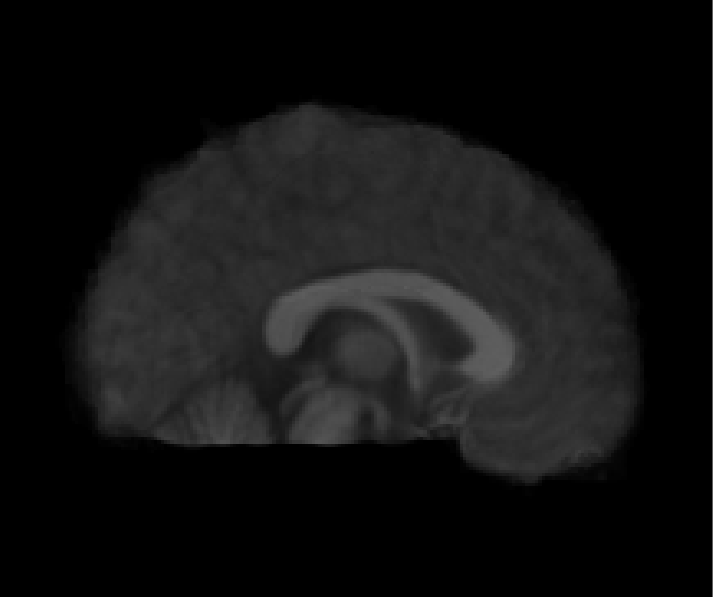}
\endminipage
\\
\quad\quad\quad\minipage{0.25\textwidth}
\vspace{0.25cm}
  \includegraphics[width=3cm]{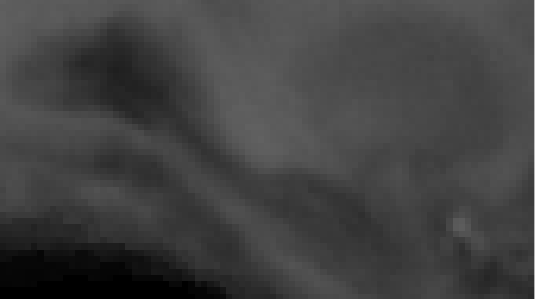}
 	\label{fig:awesome_image1}
\endminipage
\hspace{0.1cm}
\minipage{0.237\textwidth}
\vspace{0.25cm}
  \includegraphics[width=3cm]{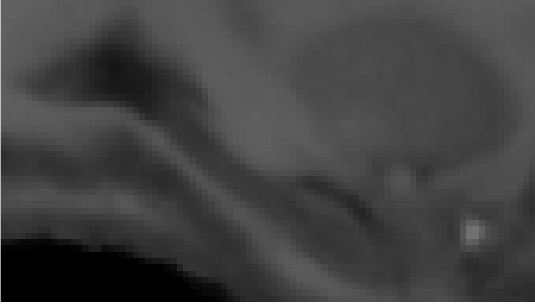}
\label{fig:awesome_image2}
\endminipage
\hspace{0.3cm}
\minipage{0.25\textwidth}%
\vspace{-0.2cm}
  \includegraphics[width=3cm]{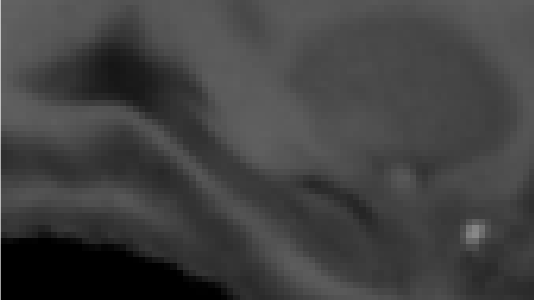}
\endminipage
\caption{Pairwise mean and target overlaps; Unaligned template and aligned template; Evolution of the appearance of hippocampus and putamen across kernel bundle scales (2 scales) with leftmost being the unaligned template.}
\label{overlaps}
\end{figure}

In this paper, we evaluate the overlaps of segmentations of images mapped to the estimated template. This experiment was chosen to illustrate the performance of the underlying registration method in the paper.  Figure~\ref{overlaps} illustrates the mean and target overlaps estimated by mapping the 10 individual images to the template space and measuring pairwise overlaps. The overlaps are comparable to what state-of-the-art registration methods have obtained on the same dataset~\cite{klein:evaluation}. The kernel bundle scales are sequentially optimized. However, if one switches to a parallel optimization across scales, a significant improvement in the overlaps may be expected. Also in the illustration is the non-aligned template image and aligned template image. The sharpness of the template is particularly visible in corpus collosum. 

To further demonstrate the effectiveness of the proposed template estimation method and benefits of multi-scale deformation, we visually inspect the atlas estimated in key regions like hippocampus and putamen. As illustrated in Figure~\ref{overlaps}, as the kernel bundle resolution becomes finer, the sharper are the boundaries of the anatomical structure. 
\section{Discussion and Conclusion}
We presented a class of non-linear mixed effect models for estimating 3D brain templates, and we proposed to simultaneously estimate both bias field and deformation parameters in a data-driven maximum likelihood setting. In addition, we proposed to incorporate the kernel bundle framework in the random deformation effects to account for deformations occurring at different scales. We illustrated the application of our in both  bias field correction and template estimation. The basis for the simultaneous estimation of the deformation and spatially correlated bias was results of~\cite{Raket:2014uf} where the authors showed that estimation of deformation parameters is biased if the data is preprocessed first followed by the prediction of deformation parameters. We modeled deformations as random geodesics on the diffeomorphism group and estimated variance parameters of both the deformation and bias field from data instead of setting them ad-hoc. In future work, we will investigate replacing random geodesics by more geometrically natural distributions on the space of diffeomorphism~\cite{Sommer:2015oh}. An important next step for this work will be to extend the method to account for multiple population means. 

{\bibfont
\bibliography{wendlandKernels}}

\begin{thebibliography}{10}

\bibitem{Joshi:2004cr}
Joshi, S., Davis, B., Jomier, M., Gerig, G.:
\newblock Unbiased diffeomorphic atlas construction for computational anatomy.
\newblock Neuroimage \textbf{23} (2004)  S151--60

\bibitem{Zhang:2013it}
Zhang, M., Singh, N., Fletcher, P.:
\newblock Bayesian estimation of regularization and atlas building in
  diffeomorphic image registration.
\newblock IPMI (2013)  37--48

\bibitem{Allassonniere:2010ya}
Allassonniere, S., Kuhn, E.:
\newblock Stochastic algorithm for parameter estimation for dense deformable
  template mixture model.
\newblock ESAIM-PS \textbf{14} (2010)  382--408

\bibitem{Raket:2014uf}
Raket, L., et~al:
\newblock A nonlinear mixed-effects model for simultaneous smoothing and
  registration of functional data.
\newblock Pattern Recognition Letters (2014)  1--7

\bibitem{Sommer:2012gu}
Sommer, S., Lauze, F., Nielsen, M., Pennec, X.:
\newblock Sparse multi-scale dieomorphic registration: the kernel bundle
  framework.
\newblock JMIV \textbf{46}(3) (2012)  292--308

\bibitem{Pai:2015di}
Pai, A., Sommer, S., Sorensen, L., Darkner, S., Sporring, J., Nielsen, M.:
\newblock Kernel bundle diffeomorphic image registration using stationary
  velocity fields and {W}endland basis functions.
\newblock IEEE TMI \textbf{PP}(99) (2015)

\bibitem{Kochunov:2001mo}
Kochunov, P., Lancaster, J., Thompson, P., Woods, R., Mazziotta, J., Hardies,
  J., Fox, P.:
\newblock Regional spatial normalization: toward an optimal target.
\newblock J Comput Assist Tomogr \textbf{25}(5) (2001)  805--816

\bibitem{Rueckert:2003ib}
Rueckert, D., et~al:
\newblock Automatic construction of 3{D} statistical deformation models of the
  brain using non-rigid registration.
\newblock IEEE TMI \textbf{22}(8) (2003)  1014--1025

\bibitem{Vialard:2011rq}
Vialard, F.X., Risser, L., Holm, D., Rueckert, D.:
\newblock Diffeomorphic atlas estimation using {K}archer mean and geodesic
  shooting on volumetric images.
\newblock MIUA (2011)

\bibitem{Zollei:2007ta}
Zollei, L., Jenkinson, M., Timoner, S., Wells, W.:
\newblock A marginalized {MAP} approach and {EM} optimization for pair-wise
  registration.
\newblock IPMI \textbf{4584}(662--674) (2007)

\bibitem{Hromatka:2015px}
Hromatka, M., et~al:
\newblock A hierarchical {B}ayesian model for multi-site diffeomorphic image
  atlases.
\newblock LNCS in AI \textbf{9350}(372-379) (2015)

\bibitem{Henderson:1950yr}
Henderson, C.R.:
\newblock Estimation of genetic parameters.
\newblock Biometrics \textbf{6} (1950)  186--187

\bibitem{Si:2014sh}
Si, S., et~al:
\newblock Memory efficient kernel approximation.
\newblock ICML (2014)

\bibitem{Robinson:1991fl}
Robinson, G.:
\newblock That {BLUP} is a good thing: The estimation of random effects.
\newblock Statistical Science \textbf{6}(1) (1991)  15--51

\bibitem{Wendland:1995hb}
Wendland, H.:
\newblock Piecewise polynomial, positive definite and compactly supported
  radial functions of minimal degree.
\newblock ACM \textbf{4}(1) (1995)  389--396

\bibitem{klein:evaluation}
Klein, A., et~al:
\newblock Evaluation of 14 nonlinear deformation algorithms applied to human
  brain {MRI} registration.
\newblock NeuroImage (2009)  786--802

\bibitem{Sommer:2015oh}
Sommer, S.:
\newblock Anisotropic distributions on manifolds: template estimation and most
  probable paths.
\newblock Information Processing in Medical Imaging (2015)  193--204

\end{thebibliography}
\bibliographystyle{splncs}
\end{document}